\documentclass[nohyperref]{article}

\usepackage{microtype}
\usepackage{graphicx}
\usepackage{subfigure}
\usepackage{booktabs} 

\usepackage{hyperref}

\usepackage[accepted]{icml2022}

\usepackage{amsmath}
\usepackage{amssymb}
\usepackage{mathtools}
\usepackage{amsthm}
\usepackage[utf8]{inputenc} 
\usepackage[T1]{fontenc}    
\usepackage{hyperref}       
\usepackage{url}            
\usepackage{booktabs}       
\usepackage{amsfonts}       
\usepackage{nicefrac}       
\usepackage{microtype}      
\usepackage{xcolor}         
\usepackage[utf8]{inputenc} 
\usepackage[T1]{fontenc}    
\usepackage{hyperref}       
\usepackage{url}            
\usepackage{booktabs}       
\usepackage{amsfonts}       
\usepackage{nicefrac}       
\usepackage{microtype}      
\usepackage{amsmath}
\usepackage{color}
\usepackage{caption}
\usepackage{subfigure}
\usepackage{graphicx}
\usepackage{multirow}
\usepackage{wrapfig}
\usepackage{url}
\usepackage{amssymb}

\usepackage[capitalize,noabbrev]{cleveref}

\theoremstyle{plain}
\newtheorem{theorem}{Theorem}[section]

\theoremstyle{definition}

\newtheorem{assumption}[theorem]{Assumption}
\theoremstyle{remark}

\usepackage[textsize=tiny]{todonotes}
\usepackage{comment}

\def\semicolon{;}
\def\applytolist#1{
    \expandafter\def\csname multi#1\endcsname##1{
        \def\multiack{##1}\ifx\multiack\semicolon
            \def\next{\relax}
        \else
            \csname #1\endcsname{##1}
            \def\next{\csname multi#1\endcsname}
        \fi
        \next}
    \csname multi#1\endcsname}
\def\calc#1{\expandafter\def\csname c#1\endcsname{{\mathcal #1}}}
\applytolist{calc}QWERTYUIOPLKJHGFDSAZXCVBNM;
\def\bbc#1{\expandafter\def\csname bb#1\endcsname{{\mathbb #1}}}
\applytolist{bbc}QWERTYUIOPLKJHGFDSAZXCVBNM;

\icmltitlerunning{Estimating Treatment Effects from Irregular Time Series Observations with Hidden Confounders}

\begin{document}

\onecolumn
\icmltitle{Estimating Treatment Effects in Continuous Time with Hidden Confounders}



\icmlsetsymbol{equal}{*}

\begin{icmlauthorlist}
\icmlauthor{Defu Cao}{equal,usc}
\icmlcorrespondingauthor{Defu Cao}{defucao@usc.edu}
\icmlauthor{James Enouen}{equal,usc}
\icmlauthor{Yan Liu}{usc}
\end{icmlauthorlist}

\icmlaffiliation{usc}{University of Southern California, Los Angeles, USA}


\icmlkeywords{Machine Learning, ICML}

\vskip 0.3in



\printAffiliationsAndNotice{\icmlEqualContribution} 

\begin{abstract}

Estimating  treatment effects plays a crucial role in causal inference, having many real-world applications like policy analysis and decision making. 
Nevertheless, estimating treatment effects in the longitudinal setting in the presence of hidden confounders remains an extremely challenging problem.
Recently, there is a growing body of work attempting to obtain unbiased ITE estimates from time-dynamic observational data by ignoring the possible existence of hidden confounders. 
Additionally, many existing works handling hidden confounders are not applicable for continuous-time settings.
In this paper, we extend the line of work focusing on deconfounding in the dynamic time setting in the presence of hidden confounders.
We leverage recent advancements in neural differential equations to build a latent factor model using a stochastic controlled differential equation and Lipschitz constrained convolutional operation in order to continuously incorporate information about ongoing interventions and irregularly sampled observations.
Experiments on both synthetic and real-world datasets highlight the promise of continuous time methods for estimating treatment effects in the presence of hidden confounders.
\end{abstract}

\section{Introduction}
\label{submission}

Estimating causal effects allows decision-makers to analyze  desired outcome from treatments and covariates accordingly, which is important in domains including marketing~\cite{brodersen2015inferring, abadie2010synthetic}, education~\cite{mandel2014offline}, healthcare~\cite{kuzmanovic2021deconfounding}, etc.
However, the existence of hidden confounders that influence treatment assignments and potential outcomes may introduce bias in the estimation \cite{simpson1951interpretation, pearl2000models}. 
Hidden confounders are not reflected in records or difficult to observe in individual treatment effect (ITE) estimation.
Take the case of cancer patients as an example:
drug resistance is a hidden confounder as it affects not only multiple treatments (procedures) but also patient outcomes (mortality, risk factors)~\cite{vlachostergios2018treatment}. 
Historical work on identifying causal effects from observational data does not focus on the challenging domain of confounders with time-varying impact.

Generally, hidden confounders can have their own time dynamics and the timing of treatments may vary between patients.
Increasingly, large-scale observational data from real world make it possible to infer causality with time-varying covariates and treatments.   \citeauthor{wang2019blessings} introduce the deconfounding framework in a static setting where they first estimate hidden confounders using a latent factor model and then infer potential outcomes using bias adjustment.
For modeling hidden confounders over time, deconfounding based models,  including TSD \cite{bica2019TSD}, SeqConf~\cite{hatt2021sequential}, and DTA~\cite{kuzmanovic2021deconfounding}, use latent variables given by their factor model as substitutes for the hidden confounders to render the assigned treatments conditionally independent. 
However, those works either do not handle irregularly-sampled, sparse, or intermittent time series or make strong assumptions (e.g., simple multivariate Gaussian distribution) for modeling the irregular samples ~\cite{bahadori2020debiasing, mastakouri2021necessary}, which make it difficult to construct the shift of causal dependency among irregular samples. Additionally, few methods use stochastic factor models designed for continuous time, resorting to architecture-based randomness like parameter dropout.

In this work, we consider the task of estimating treatment effects under continuous-time settings with multi-cause hidden confounders.
We are confronted with two major challenges: first, previous works do not consider the boundary of hidden confounders, the interval that hidden confounders can impact the relationship between treatments and outcomes; second, many practical applications involve continuous time series with irregular samples \cite{zhangcounterfactual}, which makes it difficult for existing models to make inference accurately. 
To tackle the two challenges above, we propose a Lipschitz-bounded neural stochastic controlled differential equation (LipSCDE) which uses a  stochastic differential equation factor model and Lipschitz convolutional regularization to obtain time-varying representations in continuous time with bounded influence.
We demonstrate the effectiveness of the proposed model in performing unbiased estimation, conducting multiple experiments on  both  simulated and real world datasets.
Experimental results show that LipSCDE outperforms other state-of-the-art treatment effect estimation approaches for causal inference.

We summarize the main \textbf{contributions} as follows:
(1) LipSCDE allows the continuous inclusion of input interventions and supports irregularly sampled time series by leveraging controlled differential equations and a continuous-time stochastic factor model;
(2) LipSCDE allows unbiased  estimation by utilizing Lipschitz bounded convolutional operation on high-frequency and low-frequency components from observed data to generate hidden confounders;
(3) Experiments on MIMIC III and COVID-19 demonstrate that LipSCDE achieves better estimates of individualized treatment effect and enables improved recommendations for healthcare patients.

\section{Related Work}

A large body of historical work has focused on causal estimation from observational data in the static setting.
In recent years, there has been a great increase in interest in the study of causal inference accomplished through representational learning~\cite{hill2011bayesian, yao2018representation,louizos2017causal,kallus2018causal,oezyurt2021attdmm, curth2021inductive}.
More recently, works have been moving away from the strong ignorability assumption which supposes there are no hidden confounding factors \cite{berrevoets2020organite,curth2021inductive,guo2020learning}.
\citeauthor{wang2019blessings} use a latent factor model and then infer potential outcomes using bias adjustment. Nevertheless, such works fail to handle dynamic environments.
We emphasize this work in causal inference is different from the research line of causal discovery, where proposed approaches include: \cite{spirtes2000causation, tian2013causal, peters2014causal, huang2019causal}.

In the dynamic setting with time-series data \cite{cao2020spectral,  meng2022physics}, methods for estimating treatment effects include the g-computation formula, g-estimation of structural nested mean models \cite{hernan2010causal}, and inverse probability of treatment weighting in marginal structural models (MSMs) \cite{robins2000marginal, fitzmaurice2008estimation}.
Recurrent marginal structural networks (RMSNs) \cite{lim2018forecasting} is proposed to further improve MSM's ability by capturing nonlinear dependencies.
In addition, Gaussian processes ~\cite{schulam2017reliable}, bayesian nonparametrics~\cite{roy2017bayesian} have been tailored to estimate treatment response in a continuous-time settings in order to incorporate non-deterministic quantification.
Dynamic works avoiding strong ignorability mostly follow two paths: assuming the existence of proxy variables \cite{pearl2012measurement,kuroki2014measurement,veitch2020adapting} or following the deconfounding framework \cite{bica2019TSD,hatt2021sequential,kuzmanovic2021deconfounding,ma2021deconfounding}.

\section{Problem Setup}

Here we define the problem of estimating treatment effects from irregular time series observations formally: for each patient $i\in\{1,\dots,n\}$ we have observational data at $m_i$ irregular time steps, where $t_{0}^i<\cdots<t_{m_i}^i$ where $t_{m_i}^i$ is dependent on the $i$-th patient.
We will further omit the patient id $i$ on timestamps unless they are explicitly needed. 
We have covariates $X_t^i = [x_{t_0}^i,x_{t_1}^i,  \ldots, x_{t_{m}}^i] \in \mathcal{X}_t$ and treatments $A_t^i = [a_{t_0}^i,a_{t_1}^i,  \ldots, a_{t_{m}}^i] \in \mathcal{A}_t$, where $a_{t_k}\in\bbR^j$ and $x_{t_k}\in\bbR^d$ are recorded at each timestep $t_k$.
Additionally, we have the hidden confounder variables $Z_t^i = [z_{t_0}^i,z_{t_1}^i,  \ldots, z_{t_{m}}^i] \in \mathcal{Z}_t$. 
Combining all hidden historical data with all observed historical covariates and treatments, we define the complete history $H_{< t_{m}}^i =\{X_{< t_m}^i, A_{< t_m}^i, Z_{< t_m}^i\}$ as the collection of all historical information.

We focus on one-dimensional outcomes $Y_t^i = [y_{t_0}^i,y_{t_1}^i,  \ldots, y_{t_{m}}^i] \in \mathcal{Y}_t$ and often we will be interested in the expected outcome of the future $Y^i_{a,{\geq t_k}} =  \mathbb{E}[Y^i_{a_t,t_{\geq t_k}}|H^i_t, X_{t}^i, A_{t}^i, Z_{t}^i]$ or the final expected outcome $Y^i_{a,{t_m}} =  \mathbb{E}[Y^i_{a_t,t_m}|H^i_t, X_{t}^i, A_{t}^i, Z_{t}^i]$, given a specified treatment plan $a$.
In this way, we can define the individual treatment effect (ITE) with historical data as $\tau_{a,b}^i = Y^i_{a, t_m} - Y^i_{b, t_m}$ and the average treatment effect (ATE) as $T_{a,b} = \frac{1}{n}\sum_{i=1}^{n}{\tau_{a,b}^i}$ comparing treatment plan $a$ to treatment plan $b$.
We will often be interested in the effect of a counterfactual treatment plan $a$ compared to the factual treatment $A$, which we will denote similar to before: $\tau_{a}^i = Y^i_{a, t_m} - Y^i_{t_m}$ and $T_{a} = \frac{1}{n}\sum_{i=1}^{n}{\tau_{a}^i}$.

In practice, we rely on assumptions to be able to estimate the individual treatment effect.
We leverage the standard causal assumptions of Positivity/ Overlap~\cite{imai2004causal} and Consistency~\cite{lim2018forecasting} which say that all possible treatment plans occur with positive probability and that enforcing a treatment plan has the same result as such a treatment sequence naturally occurring.
We also make Assumption ~\ref{ass:sssi} below which aligns our work with other dynamic-time works in the deconfounding setting.
Assumption ~\ref{ass:sssi} expands the sequential single strong ignorability assumption from ~\cite{bica2019TSD} to the continuous-time setting, so multi-cause hidden confounders exist at every time and have a causal effect on the treatment $A_t$ and potential outcome $Y_t$.
\begin{assumption}
{
\label{ass:sssi}
\textit{Sequential single strong ignorability in continuous-time setting.} 
If there exists multi-cause confounders, we have
$Y_{a_{\geq t_m}} \perp \!\!\! \perp  A_{{t_m},j}|X_{t_m}, H_{< t_{m}}$, for all $a_{\geq t_m}$ and all
$j$ possible assigned treatments.
}
\end{assumption}

\begin{figure*}[t]
\centering
\includegraphics[width=\linewidth]{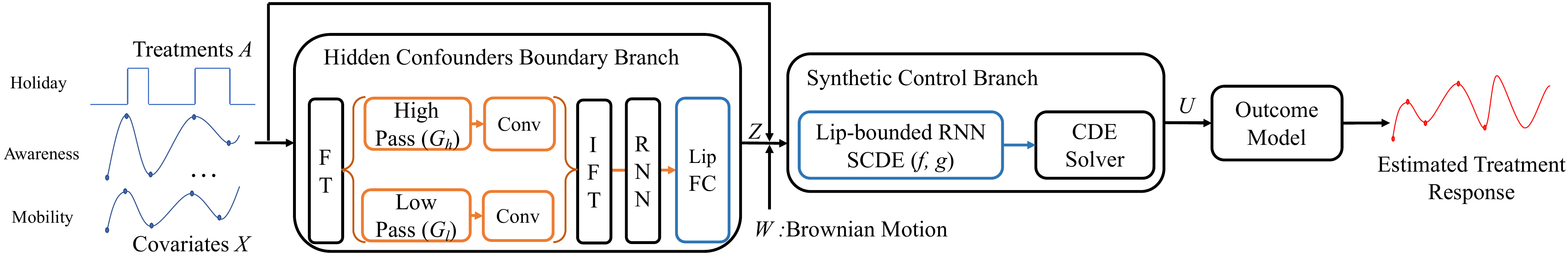}
\caption{Architecture of LipSCDE.}
\vspace{-0.5cm}
\label{fig:network}
\end{figure*}

\section{Lipschitz Bounded Neural Stochastic Controlled Differential Equations (LipSCDE)}

To address the  treatment effects estimation task from irregular time series observation, we must avoid the inference bias caused by hidden confounders. 
Thus, we propose an approach called Lipschitz-bounded neural stochastic controlled differential equations (LipSCDE). 
Figure~\ref{fig:network} illustrates the network architecture of LipSCDE, which first infers the interrelationship of hidden confounders on treatment by bounding the boundary of hidden confounders, and then models the history trajectories with neural network parameterized by a stochastic differential equation factor model (SDE) and a Lipschitz controlled differential equation (CDE).

\textbf{Hidden Confounders Boundary Branch.}
This section focuses on how to use Lipschitz regularized convolutional operation to infer the hidden confounders from both high-frequency signals and low-frequency signals of observed data. 
Generally,  high frequency components correspond to the fast transformed parts in time domain which corresponds to  boundaries of the data that are susceptible to change, while low frequency components contain smoothing information which contains the trend of observed data. 
We find that our Lipschitz regularization allows us to balance the bounded influence of hidden confounders with the dynamic changes in observed covariates.

We convert our history trajectories $h_t$ from the time-domain to the frequency-domain using the Fourier transform $\cF$.
We next use Gaussian high-pass filter $G_{h}$ and Gaussian low-pass filter $G_{l}$ to get high frequency and low frequency components.
We then use Lipschitz constrained convolutional operators and apply the inverse Fourier transform $\mathcal{F}^{-1}$.
Afterwards, we apply a small, Lipschitz-regularized multilayer perceptron \cite{perugachidiaz2021invertible,gouk2021regularisation}. 
This representation is then passed to the next phase of the model.

\textbf{Synthetic Control Branch.}
Since the neural ordinary differential equations (ODE) family is effective in continuous time series problems, we use a combination of neural CDE and neural SDE to estimate latent factors and treatment effects. 
Neural differential equations allow us to estimate the nonlinear influence in continuous time and gradient computation is enabled by the adjoint method~\cite{chen2018neural}.

Let $u_{t}:=h_{\theta}\left(H_{t}\right)= h_{\theta} \left([x_t,a_t,z_t, H_{< t}]\right)$, where $h_{\theta}$ embeds the observations into a $l$-dimensional latent state.
Let $f_\theta$ and $g_\theta$ be neural networks parameterizing the latent vector fields.
We follow the equation ~\cite{erichson2020Lipschitz} to define $f_\theta$ as a continuous-time Lipschitz RNN, parameterized by $\theta$:
$
f_\theta(h,t)=A_\theta h+ \sigma \left(W_\theta h+U_\theta u_t+b\right),
$
where hidden-to-hidden matrices $A_{\theta}$, $W_\theta$, and $U_\theta$ are trainable matrices and nonlinearity $\sigma(\cdot)$ is a 1-Lipschitz function.  We define $g$ similarly.
A latent path can then be expressed as the solution to the stochastic controlled differential equation of the form:
\begin{align}
u_{t}=u_{t_{0}}+\int_{t_{0}}^{t} f_\theta\left(u_{s}, s\right) d \mathbf{H}_{s}+\int_{t_{0}}^{t} g_\theta\left(u_{s}, s\right) d \mathbf{W}_{s}, \quad t \in\left(t_{0}, t_{m}\right]
\label{eqn:csde}
\end{align}

These differential equations are controlled by the historical information $\mathbf{H}$ and the stochastic Brownian motion $\mathbf{W}$.
For each estimate of $f_{\theta}$ and $g_{\theta}$, the forward latent trajectory that these functions define through (\ref{eqn:csde}) can be computed using any numerical ODE solver, and the gradients can be computed leveraging recent advances in adjoint-based differentiable training \cite{kidger2020neuripsNeuralCDEs,kidger2021icmlNeuralSDEsInfiniteGANs}.

\textbf{Outcome Model.}
After sampling the latent representation $U_t = (\hat{u}_{t_{1}}, \ldots, \hat{u}_{t_{k}})$ of history trajectories on each individual, we use the outcome model to estimate unbiased treatment effect on the outcome response via inverse probability of treatment weighting (IPTW)~\cite{lim2018forecasting}:
$\mathbb{E}[Y_{a,{\geq t}} | U_t]$.  
Specifically, we use two stacked LSTM CDE layers to decode the padded hidden sequence  of irregular inputs. Then we use a linear, fully-connected layer to map the output into the final outcomes.

\section{Experiments}

\textbf{Experiments Settings.}
In this section, we estimate the treatment effects for each time step by one-step ahead predictions on both synthetic dataset and real world datasets including MIMIC-III~\cite{johnson2016mimic} dataset and COVID-19~\cite{steiger2020causal} dataset. We adopt the Adam~\cite{kingma2014adam} optimizer with learning rate 0.01. The training epoch is set as 10 with 10 iterations on each batch. The batch size is 16 and each dataset follows a 80/10/10 split for training/validation/testing respectively. Except LipSCDE, all baselines share the same design of the outcome model, i.e., \textbf{MSM}~\cite{robins2000marginal}, which uses  IPTW to adjust for the time-dependent confounding bias by linear regression and then constructs a pseudo-population to compute final outcome, and \textbf{RMSN}~\cite{lim2018forecasting}, which estimates IPTW using RNNs instead of logistic regressions.

\begin{wraptable}{l}{0.5\textwidth}
\small
\centering
\setlength{\abovecaptionskip}{0.2cm}  
\setlength{\belowcaptionskip}{-0.5cm}   
\caption{Results for real-world data (MIMIC III and COVID-19) experiments. Lower is better.}
\begin{tabular}{ccccc}
\hline \hline Outcome  &\multirow{2}{*}{Method} &\multicolumn{3}{c}{ RMSE (\%) } \\
\cline { 3 - 5} model&& Blo. pre. & Oxy. sat. & COVID-19 \\
\hline \multirow{3}{*}{MSM}&Conf. & $14.54 $ & $4.72 $ & $15.10$  \\
&DTA & $13.31 $ & $4.65 $ & $13.93 $  \\
&TSD & $13.57 $ & $4.33 $ & $13.07 $  \\
\hline \hline
\multirow{3}{*}{RMSN}&Conf. & $14.46 $ & $4.22 $ & $11.48 $  \\
&DTA & $18.33 $ & $4.21 $ & $13.52$  \\
&TSD & $12.11 $ & $4.25$ & $11.08 $  \\
&SeqDec & $13.74 $ & $4.19$ & $11.43 $  \\
\hline \hline
Ours &LipSCDE & \textbf{8.82 } & \textbf{4.19} & \textbf{10.28 }  \\
\hline \hline
\end{tabular}
\vspace{-0.6cm}
\label{tab:mimic}
\end{wraptable}

\textbf{Baselines.} LipSCDE is evaluated by the degree of control it has over hidden confounders. We thus compare against suitable baselines which are designed for ITE estimation in a dynamic-time settings controlling for hidden confounders. The baselines used in this experiments are: 
\textbf{Conf.}, which ignores the existence of hidden confounder; \textbf{TSD}~\cite{bica2019TSD}, which leverages the presence of multi-cause hidden confounders; \textbf{DTA}~\cite{kuzmanovic2021deconfounding}, which combines a long short-term memory autoencoder with a causal regularization penalty to learn a hidden embedding; \textbf{SeqDec}~\cite{hatt2021sequential}, which utilizes a Gaussian process latent model to infer confounder substitutes.

\begin{wraptable}{r}{0.5\textwidth}
\small
\centering
\vspace{-0.3cm}
\caption{Irregular data with missing rate  of \{0\%, 15\%, 30\%\}.}
\begin{tabular}{lccccc}
\hline\hline
M/R                                   & Degree & Conf. & TSD & SeqDec & LipSCDE \\ \hline 
\multirow{3}{*}{0\%}                     & 0      & 2.5   & 2.73 & 2.17   & \textbf{0.60}   \\
                                          & 0.2    & 3.01   & 2.66 & 2.32   & \textbf{1.74}    \\
                                          & 0.4    & 3.83   & 3.37 & 3.70    & \textbf{2.17}   \\ \hline \hline 
\multirow{3}{*}{15\%}                     & 0      & 3.43   & 2.83 & 2.43   & \textbf{1.64}   \\
                                          & 0.2    & 3.47   & 2.84 & 2.69   & \textbf{1.56}    \\
                                          & 0.4    & 3.45   & 3.67 & 3.70    & \textbf{2.65}   \\ \hline \hline 
\multirow{3}{*}{30\%} & 0      & 3.32   & 2.84 & 3.19   & \textbf{2.21}   \\
\multicolumn{1}{c}{}                      & 0.2    & 4.66   & 3.65 & 2.95   & \textbf{2.81}  \\
\multicolumn{1}{c}{}                      & 0.4    & 4.19   & 4.06 & 3.89   & \textbf{2.96}   \\ \hline\hline
\end{tabular}
\label{tab:irr}
\vspace{-0.3cm}
\end{wraptable}

\textbf{Results.} We emphasize that our model is suitable for irregular time series sampling by randomly removing 15\% and 30\% of the aligned synthetic data independently for each unit, using confounding degrees of {0.0, 0.2, 0.4}.
Except CDE-based methods, all the baselines require some form of a prior interpolation and are evaluated on a regular grid of time points. 
As shown in Table~\ref{tab:irr}, methods considering hidden confounders are  generally better than the models without the hidden confounders. Note that, LipSCDE achieves better results on all different levels of confounders which demonstrates that our model has relatively good performance with irregularly aligned data.
Our method also outperforms existing baselines across multiple real-world tasks in Table~\ref{tab:mimic}, which is consistent with what we have seen in the simulated dataset. Specifically,
our method outperforms the TSD baseline by 27.2\% in blood pressure prediction and 7.2\% in covid case prediction. 

\section{Conclusion} 
In this paper, we proposed the Lipschitz-bounded neural stochastic controlled differential equation (LipSCDE), a novel neural network that utilizes hidden confounders for the estimate treatment effect in the case of irregular time series observations. 
It uses the time-varying observations in the frequency domain to infer the hidden confounders under Lipschitz constraints.  
Moreover, the combination of SDE and CDE explicitly models the latent path of observed time series, which can effectively capture underlying temporal dynamics and intervention effects. 
Experimentally, we show the effectiveness of LipSCDE in eliminating bias in estimating  treatment effects.
LipSCDE has potential implications for practice since it can provide powerful support for decision making like enabling better individualized treatment recommendations in healthcare.

\nocite{langley00}

\bibliography{example_paper}
\bibliographystyle{icml2022}


\end{document}